\documentclass[10pt,twocolumn,letterpaper]{article}

\usepackage{iccv}
\usepackage{times}
\usepackage{epsfig}
\usepackage{graphicx}
\usepackage{amsmath}
\usepackage{amssymb}

\usepackage{multirow}
\usepackage{booktabs}
\usepackage{diagbox}
\usepackage{array}

\usepackage{algorithm}
\usepackage{algorithmic}

\usepackage[breaklinks=true,bookmarks=false]{hyperref}

\iccvfinalcopy 

\def\iccvPaperID{7465} 

\ificcvfinal\fi

\begin{document}

\title{Adversarial Catoptric Light: An Effective, Stealthy and Robust Physical-World Attack to DNNs}

\author{Chengyin Hu\\
University of Electronic Science \\and Technology of China\\
Chengdu, China\\
{\tt\small cyhuuestc@gmail.com}
\and
Weiwen Shi\\
University of Electronic Science \\and Technology of China\\
Chengdu, China\\
{\tt\small Weiwen\_shi@foxmail.com}
\and
}

\maketitle
\ificcvfinal\thispagestyle{empty}\fi

\begin{abstract}
Deep neural networks (DNNs) have demonstrated exceptional success across various tasks, underscoring the need to evaluate the robustness of advanced DNNs. However, traditional methods using stickers as physical perturbations to deceive classifiers present challenges in achieving stealthiness and suffer from printing loss. Recent advancements in physical attacks have utilized light beams such as lasers and projectors to perform attacks, where the optical patterns generated are artificial rather than natural. In this study, we introduce a novel physical attack, adversarial catoptric light (\textbf{AdvCL}), where adversarial perturbations are generated using a common natural phenomenon, catoptric light, to achieve stealthy and naturalistic adversarial attacks against advanced DNNs in a black-box setting. We evaluate the proposed method in three aspects: effectiveness, stealthiness, and robustness. Quantitative results obtained in simulated environments demonstrate the effectiveness of the proposed method, and in physical scenarios, we achieve an attack success rate of 83.5\%, surpassing the baseline. We use common catoptric light as a perturbation to enhance the stealthiness of the method and make physical samples appear more natural. Robustness is validated by successfully attacking advanced and robust DNNs with a success rate over 80\% in all cases. Additionally, we discuss defense strategy against AdvCL and put forward some light-based physical attacks.

\end{abstract}


\section{Introduction}
\label{sec1}

Deep neural networks have made significant advances in image classification and object detection in recent years. At the same time, many vision-based applications, such as UAVs and autonomous driving, are gaining popularity. Recent advances, however, have shown that advanced DNNs are vulnerable to minor perturbations, even if only one pixel is modified \cite{ref20}. It is critical to assess the dependability of advanced DNNs in safety-critical scenarios (medical, autonomous driving, etc.). Until now, most research has focused on attacks in digital environments \cite{ref18,ref19,ref21}, which target advanced DNNs by adding imperceptible perturbations to images. Some researchers have gradually devoted themselves to the study of physical attacks \cite{ref22,ref23,ref24}, which differ from digital attacks in that images are captured by cameras and then fed to the target model. Since it is difficult to capture subtle perturbations with the camera, the physical perturbations are designed to be much larger. Physical perturbations are thus detectable by human observers. When an attacker executes physical attacks, there is a trade-off between robustness and stealthiness.

\begin{figure}
\centering
\includegraphics[width=1\columnwidth]{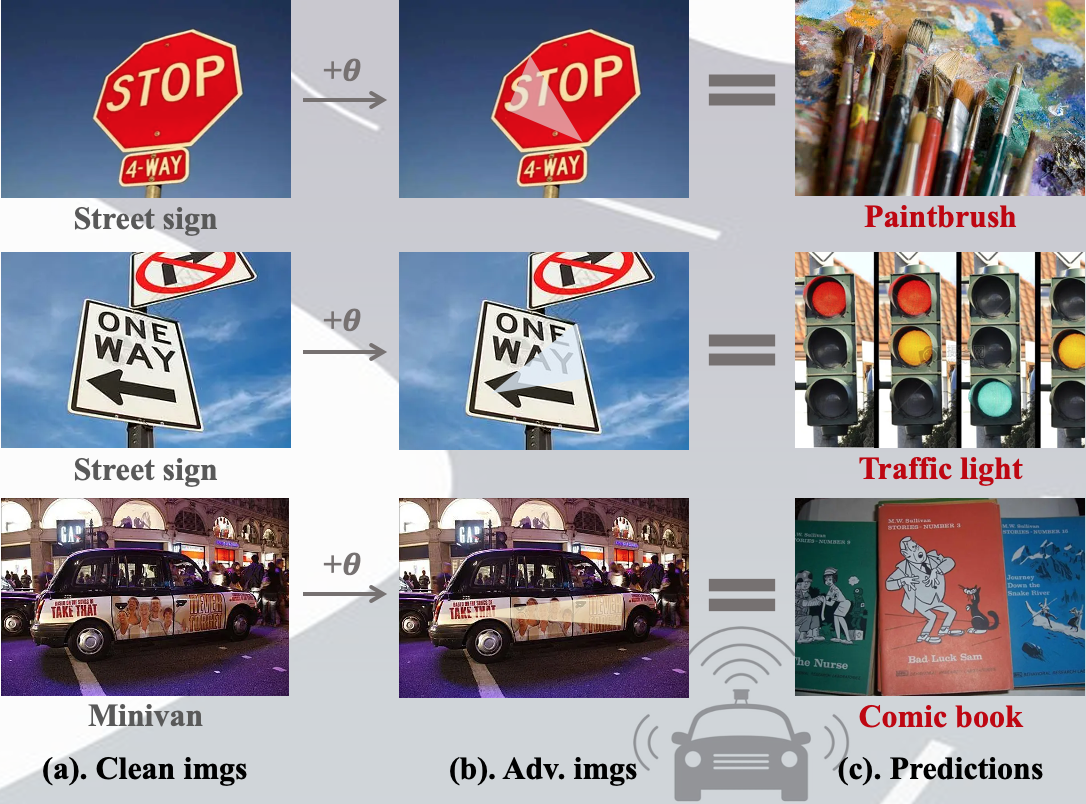} 
\caption{An example. When the camera of a self-driving car captures objects projected with carefully designed catoptric light, it fails to recognize the "Street sign" and "Minivan".}.
\vspace{-5mm}
\label{figure1}
\end{figure}

Many natural phenomena, such as natural light and shadows, play the role of physical perturbations, which can lead to tragic crashes involving self-driving cars. If common catoptric light is used as an attack weapon to launch physical attacks against advanced DNNs, vision-based systems may be jeopardized. At the same time, given the common natural catoptric light, it may reduce defender vigilance. As shown in Figure \ref{figure1}, the attacker projects a carefully designed catoptric light onto the street sign, preventing the self-driving car from correctly recognizing it.

Until now, most physical attacks have been carried out using stickers \cite{ref24,ref26}, and such methods can usually achieve robust adversarial effects without changing the original information of the target object. However, sticker-based physical attacks are difficult to conceal. Light beams are used as physical perturbations in some physical attacks to perform physical attacks \cite{ref33,ref34,ref35,ref49}, light transience is used in such attacks to achieve physical stealthiness. These light-based perturbation patterns, on the other hand, are man-made rather than natural. \cite{ref37} performs attacks using common shadows as natural adversarial perturbations, which studies the effects of daytime attacks. Our method, on the other hand, studies physical attacks at night by employing catoptric light as natural adversarial perturbations. Furthermore, some researchers have investigated camera-based physical attacks \cite{ref38}, in which a tiny translucent patch is stuck to a mobile phone's camera to perform physical attacks. However, the physical samples generated by such an attack can easily raise defenders' suspicions.

\begin{figure}
\centering
\includegraphics[width=1\columnwidth]{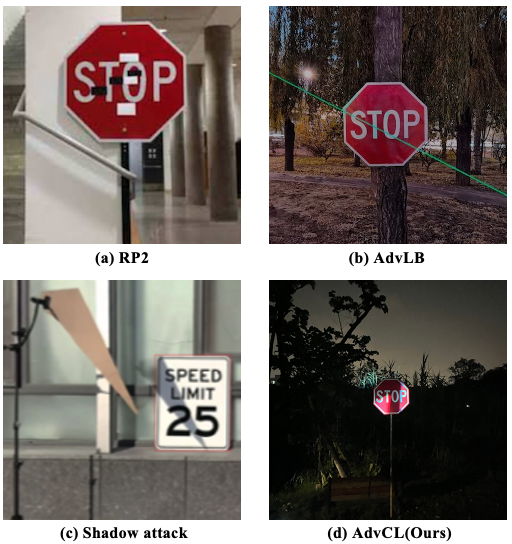} 
\caption{Visual comparison.}.
\vspace{-5mm}
\label{figure2}
\end{figure}

We present a novel light-based physical attack called adversarial catoptric light (AdvCL) in this paper. Unlike traditional sticker-based physical attacks, our approach uses light as a perturbation, giving AdvCL flexibility. In contrast to recent light-based physical attacks, our approach employs a common phenomenon, catoptric light, as perturbations, making our physical sample appear more natural. Figure \ref{figure2} shows a visual comparison of physical samples created using AdvCL and other methods. The adversarial sample generated by AdvCL is stealthier than other methods such as RP2 \cite{ref24}, AdvLB \cite{ref35}, and shadow attack \cite{ref37}.

Our approach is easy to deploy physical attacks, by formalizing the physical parameters of catoptric light, genetic algorithm \cite{ref46} is used to find the most aggressive ones, then based on which, projecting catoptric light to the target objects to generate physical samples. Notably, our approach enables low-cost attacks, requiring only a budget of less than 20 USD, rendering it significantly easier to deploy.
Our main contributions are summarized as follows:

\begin{itemize}
\item We propose a new light-based attack called adversarial catoptric light (AdvCL), which takes advantage of the properties of light to launch an effective, stealthy, and robust attack (See Section \ref{sec1}). At the same time, the total cost of our devices is under 20 USD, making it simple to launch such an attack.

\item We present and analyze existing methods (See Section \ref{sec2}), demonstrate effective optimization strategies, and conduct extensive experiments to validate AdvCL's effectiveness, stealthiness, and robustness (See Section \ref{sec3}, Section \ref{sec4}). Our results indicate that AdvCL is capable of achieving high success rates while remaining inconspicuous to the naked eye, even in challenging real-world scenarios. Given these findings, we believe that AdvCL represents a valuable tool for further studying the threat posed by light-based attacks in realistic settings.

\item We conduct adversarial defense against AdvCL via adversarial training (See Section \ref{sec5}). At the same time, we propose some new light-based physical attacks (See Section \ref{sec6}).

\end{itemize}

\begin{figure*}
\centering
\includegraphics[width=1\linewidth]{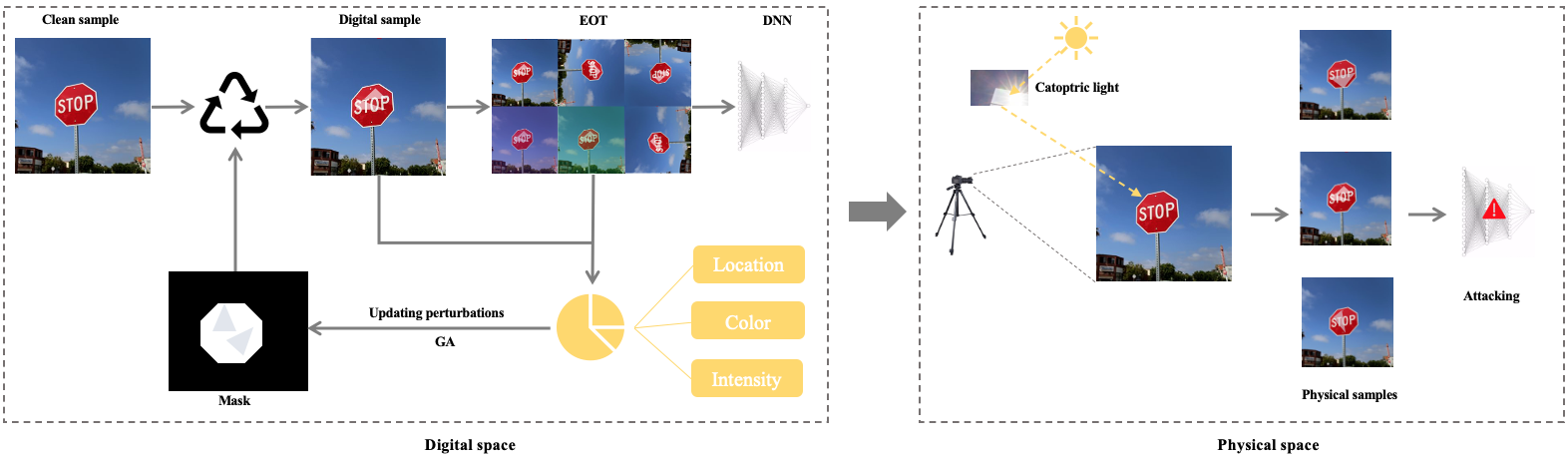} 
\caption{The attacker uses genetic algorithm to optimize the simulation samples, and uses EOT to achieve the domain transition of digital samples to physical samples. Finally, the catoptric light is projected onto the target object to fool the classifier.}.
\vspace{-2mm}
\label{figure3}
\end{figure*}


\section{Related work}
\label{sec2}
\subsection{Digital attacks}

Szegedy et al. \cite{ref1} proposed the adversarial attack first, demonstrating that advanced DNNs are susceptible to interference from minor perturbations, after which adversarial attacks were successfully proposed \cite{ref16,ref17,ref20,ref21}.

Most digital attacks keep adversarial perturbations small enough to be imperceptible to human observers. Among them, the most commonly used norms are ${l}_{2}$ and ${l}_{\infty}$ \cite{ref2,ref3,ref4,ref5,ref6}. Furthermore, some researchers alter other attributes of digital images, such as color \cite{ref7,ref8,ref9}, texture and camouflage \cite{ref10,ref11,ref12,ref13}, to generate adversarial samples. These methods produce perturbations that are barely detectable by human observers. At the same time, some researchers alter the physical parameters of digital images \cite{ref14,ref15}, retaining only the key components in order to generate adversarial samples. In general, digital attacks assume that an attacker can modify the input image, but this is not practical in a physical scenario.

\subsection{Physical attacks}

Kurakin et al. \cite{ref22} were the first to propose physical attack. Following this work, many physical attacks were proposed\cite{ref24,ref28,ref29,ref30,ref31}.

\textbf{Traditional street sign attacks.} Ivan Evtimov et al. proposed RP2 \cite{ref24}, which performs attacks against advanced DNNs by using black-white stickers as perturbations. However, at long distances and angles, RP2 is susceptible to environmental interference. Eykholt et al. \cite{ref26} improved RP2 to generate robust and transferable adversarial samples in order to fool advanced DNNs. However, the perturbations cover a large enough area to be noticeable. Chen et al. \cite{ref23} and Huang et al. \cite{ref27} proposed ShapeShifter and the improved ShapeShifter, which perturb areas beyond the "stop" and were successful in fooling advanced DNNs. However, they have a flaw: the perturbations cover almost the entire road sign, preventing it from being stealthy. AdvCam, proposed by Duan et al. \cite{ref25}, generates adversarial samples and disguises perturbations as a style considered reasonable by human observers, but it requires manual selection of the attack area and target. Overall, the methods described above necessitate manual modification of the target objects. Furthermore, these works fell short of achieving stealthiness.


\textbf{Light-based attack.} Duan et al. \cite{ref35} proposed AdvLB, which executes attacks by using laser beams as perturbations. It is more flexible than conventional street sign attacks. However, in physical attack scenarios, it is prone to space errors.
Gnanasambandam et al. \cite{ref36} proposed OPAD, which amplifies and projects subtle digital perturbations onto the target object to generate physical samples. Human observers, on the other hand, are suspicious of its irregular projection patterns.
Zhong et al. \cite{ref37} studied a shadow-based physical attack in which adversarial samples were generated by casting carefully crafted shadows on the target object, resulting in a natural black-box attack. This shadow attack, however, cannot be used in low-light conditions. Importantly, when carrying out such an attack, the cardboard must be placed very close to the target object (see Figure \ref{figure2}), which is easily suspicious to human observers. Our proposed AdvCL, on the other hand, uses the retroreflector as a tool and places it at a relatively large spatial distance from the target object, which raises less suspicion. Because the proposed AdvCL uses natural catoptric light as a perturbation to perform physical attacks in low-light conditions, it is a complement to shadow attack.

\textbf{Camera-based attack.} Li et al. \cite{ref38} investigated camera-based attacks by placing well-designed stickers on the camera lens to generate adversarial samples, this method avoids physically manipulating the target by physically manipulating the camera itself. However, the physical samples generated by this method are suspect, and access to the victim's camera is impractical.


\section{Approach}
\label{sec3}
Given an input picture $X$, a ground truth label $Y$, and a DNN classifier $f$. $f(X)$ represents the classifier's prediction label for picture $X$, The classifier $f$ associates with a confidence score  ${f}_{Y}(X)$ to class $Y$. The adversarial sample ${X}_{adv}$ has two properties: (1) $f({X}_{adv}) \neq f(X) = Y$; (2)$\parallel{X}_{adv}-X \parallel < \epsilon$. Among them, the first property requires ${X}_{adv}$ fools DNN classifier $f$. The second property requires that the perturbations of ${X}_{adv}$ be small enough for human observers to notice.

In this work, we use genetic algorithm \cite{ref46} to optimize the physical parameters of catoptric light. Then, in the real scenarios, we project catoptric light onto a target object and generate physical samples. Figure \ref{figure3} shows our approach.

\begin{figure}
\centering
\includegraphics[width=1\columnwidth]{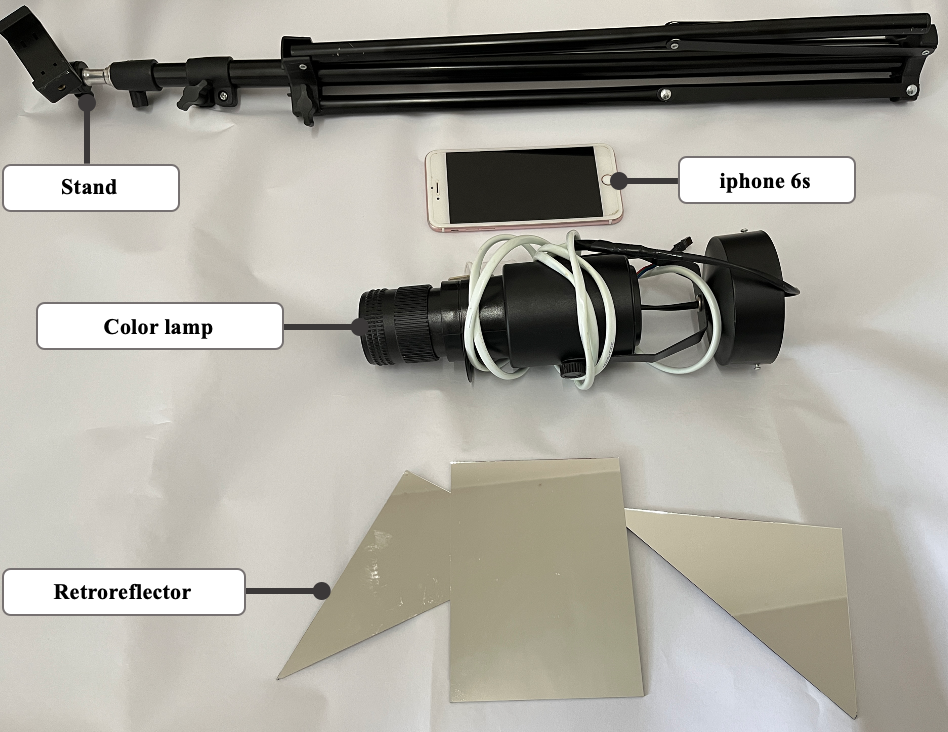} 
\caption{Experimental devices.}.
\vspace{-5mm}
\label{figure4}
\end{figure}

\subsection{Generate adverarial sample}
In this paper, we define catoptric light using three physical parameters: location ${\mathcal{P}}_{l}$, color $\mathcal{C}(r, g, b)$, intensity $\mathcal{I}$. Each parameter is described as follows:

\textbf{Location ${\mathcal{P}}_{l}$:} We propose finding a polygon ${\mathcal{P}}_{l}$ to simulate the catoptric light, which is expressed by a set of vertices $l={({m}_{1},{n}_{1}),({m}_{2},{n}_{2}),…,({m}_{k},{n}_{k})}$, for the location of catoptric light. The catoptric light region can be expressed as $\mathcal{M} \cap {\mathcal{P}}_{l}$ where $\mathcal{M}$ is used as the mask to locate the target object. Although ${\mathcal{P}}_{l}$ can be any polygon, we perform attacks with various polygons in Section \ref{sec4}, and the experimental results show that triangles are sufficient to generate successful adversarial samples. We can use various polygons to increase attack success rates, but such a catoptric light would look unnatural and would be difficult to implement in a physical scenario. As a result, in the physical experiments, we employ triangle catoptric light to carry out attacks.

\textbf{Color $\mathcal{C}(r, g, b)$:}  $\mathcal{C}(r, g, b)$ denotes the color of the catoptric light, where $r$, $g$, and $b$ denote the red, green, and blue channels of the catoptric light respectively.

\textbf{Intensity $\mathcal{I}$:} $\mathcal{I}$ indicates the intensity of catoptric light. 

Note that it is not practical to generate various colors of  catoptric light (e.g. black) in the physical environment, therefore, color is specified in the physical test for optimization (such as white, yellow). Meanwhile, we select the catoptric light intensity in the physical test using the experimental results in Table \ref{Table 1}, see Section \ref{sec4} for details.

The parameters ${\mathcal{P}}_{l}$, $\mathcal{C}(r, g, b)$ and $\mathcal{I}$ form a catoptric light’s physical parameter $\theta(\mathcal{C}, {\mathcal{P}}_{l}, \mathcal{I})$. We define a simple function $S(X; \theta(\mathcal{C}, {\mathcal{P}}_{l}, \mathcal{I}), \mathcal{M})$ that simply synthesizes the input image with catoptric light to generate an adversarial sample (for more information on how to generate adversarial sample using equation \ref{Formula 1}, see the code in the supplementary materials.), which can be expressed as follows:

\begin{equation}
    \label{Formula 1}
    {X}_{adv} = S(X; \theta(\mathcal{C}, {\mathcal{P}}_{l}, \mathcal{I}), \mathcal{M}) 
\end{equation}

\textbf{Expectation Over Transformation.} EOT \cite{ref31} is a powerful tool for managing the transition from digital to physical domains. we define a transformation $\mathcal{T}$, which is a random combination of digital image processing, such as brightness adaptation, position offset, color variation, and so on. Through EOT, the physical sample can be represented as follows:

\begin{equation}
    \label{Formula 2}
    {X}_{phy} = \mathcal{T}({X}_{adv}; \mathcal{C}, {\mathcal{P}}_{l}, \mathcal{I}) 
\end{equation}

\begin{table*}[ht]
\centering
\caption{\label{Table 1}Quantitative results at varying setups (ASR (\%)).}
\begin{tabular}{cccccccccccc}
\hline

\multicolumn{1}{c}{\multirow{2}{*}{num of edges ${\mathcal{P}}_{l}$}} & \multicolumn{1}{c}{\multirow{2}{*}{Method}} & \multicolumn{9}{c}{$\mathcal{I}$}\\
\cmidrule(r){3-11}

\multicolumn{1}{c}{} & \multicolumn{1}{c}{} & $0.1$ & $0.2$ & $0.3$ & $0.4$ & $0.5$ & $0.6$ & $0.7$ & $0.8$ & $0.9$ &
\multicolumn{1}{c}{\multirow{1}{*}{average} }\\

\hline
\multicolumn{1}{c}{\multirow{2}*{3}} & \multicolumn{1}{c}{Random} & 12.5 & 24.5 & 45.7 & 53.2 & 71.2 & 79.9 & 87.3 & 90.2 & 92.8 & 61.9  \\
\multicolumn{1}{c}{~} & \multicolumn{1}{c}{AdvCL} & 22.6 & 41.9 & 57.6 & 69.8 & 69.8 & 87.9 & 91.1 & 94.9 & 96.8 & \textbf{70.3} \\
\hline

\multicolumn{1}{c}{\multirow{2}*{4}} & \multicolumn{1}{c}{Random} & 14.6 & 26.7 & 49.9 & 58.3 & 77.6 & 82.7 & 89.6 & 92.1 & 93.8 & 65.0  \\
\multicolumn{1}{c}{~} & \multicolumn{1}{c}{AdvCL} & 25.5 & 46.4 & 60.2 & 75.7 & 86.4 & 90.3 & 92.8 & 95.1 & 97.0 & \textbf{74.4} \\
\hline

\multicolumn{1}{c}{\multirow{2}*{5}} & \multicolumn{1}{c}{Random} & 15.1 & 27.1 & 51.5 & 63.3 & 79.2 & 85.9 & 90.1 & 93.4 & 94.2 & 66.6  \\
\multicolumn{1}{c}{~} & \multicolumn{1}{c}{AdvCL} & 29.5 & 49.1 & 65.8 & 79.6 & 88.7 & 92.1 & 93.9 & 95.8 & 97.5 & \textbf{76.9} \\
\hline

\multicolumn{1}{c}{\multirow{2}*{6}} & \multicolumn{1}{c}{Random} & 17.2 & 30.3 & 53.3 & 65.1 & 82.5 & 87.3 & 92.4 & 94.1 & 95.6 & 68.6  \\
\multicolumn{1}{c}{~} & \multicolumn{1}{c}{AdvCL} & 33.9 & 53.2 & 69.7 & 83.8 & 90.6 & 93.2 & 95.1 & 96.4 & 97.9 & \textbf{79.3} \\
\hline

\multicolumn{1}{c}{\multirow{2}*{7}} & \multicolumn{1}{c}{Random} & 20.1 & 33.9 & 56.6 & 69.3 & 85.5 & 89.7 & 93.1 & 95.4 & 96.1 & 71.1  \\
\multicolumn{1}{c}{~} & \multicolumn{1}{c}{AdvCL} & 35.1 & 55.3 & 72.5 & 87.9 & 93.6 & 94.2 & 96.1 & 97.5 & 98.4 & \textbf{81.2} \\
\hline

\multicolumn{1}{c}{\multirow{2}*{8}} & \multicolumn{1}{c}{Random} & 21.3 & 35.7 & 57.9 & 71 & 86.3 & 90.6 & 95 & 96.1 & 96.7 & 72.3  \\
\multicolumn{1}{c}{~} & \multicolumn{1}{c}{AdvCL} & 36.7 & 57.5 & 74.9 & 90.2 & 94.5 & 95.7 & 96.4 & 97.9 & 98.8 & \textbf{82.5} \\
\hline

\multicolumn{1}{c}{\multirow{2}*{9}} & \multicolumn{1}{c}{Random} & 22.2 & 36.9 & 59.1 & 72.3 & 87.9 & 91.9 & 96.3 & 96.7 & 96.9 & 73.4  \\
\multicolumn{1}{c}{~} & \multicolumn{1}{c}{AdvCL} & 69.8 & 58.4 & 76.3 & 92.8 & 95.1 & 95.8 & 97.6 & 98.1 & 98.9 & \textbf{83.5} \\
\hline


\end{tabular}
\end{table*}


\subsection{Genetic algorithm}

John Holland created the genetic algorithm (GA) \cite{ref46} as a natural metaheuristic algorithm based on the laws of biological evolution in nature. It is a computational model that simulates the biological evolution process of natural selection and the genetic mechanism of Darwin's biological evolution in order to find the best solution by simulating the natural evolution process.

We use no model gradient information in this work, only the confidence score and prediction label from model feedback. Among the advantages of using GA to optimize AdvCL are:

(1) GA searches the string set of solutions of the problem, which covers a large area and is conducive to global optimization. In our method, physical parameters $\mathcal{C}$, $\mathcal{P}_{l}$ and $\mathcal{I}$ include a total of $256\times256\times256\times512\times512\times6\times4$ combinations of problem solutions (in which $\mathcal{C}$($256\times256\times256$), $\mathcal{P}_{l}$ ($512\times512\times6$), $\mathcal{I}$ (4)), GA is conducive to the global optimization of AdvCL.

(2) GA does not require any knowledge of the search space or any other auxiliary information, it uses the fitness value to evaluate individuals and performs genetic operations on this basis. The fitness function is not constrained by continuous differentiability, and the domain of its definition can be set arbitrarily. AdvCL does not require gradient information from the model and instead uses the model's confidence score ${f}_{Y}(X)$ as the individual fitness and $f({X}_{adv}) \neq Y$ as the termination condition.

(3) A flexible selection strategy. GA organizes search using evolutionary information. Individuals with high fitness have a higher chance of survival and a more adaptable gene structure. To further broaden the search scope and achieve global optimization, AdvCL employs the flexibility of GA to select specific elimination strategies.

More information on genetic algorithm optimization can be found in the supplementary material.


\begin{algorithm}
	\renewcommand{\algorithmicrequire}{\textbf{Input:}}
	\renewcommand{\algorithmicensure}{\textbf{Output:}}
	\caption{Pseudocode of AdvCL}
	\label{algorithm1}
	\begin{algorithmic}[1]
	
		\REQUIRE Input $X$, Classifier $f$, Ground truth label $Y$, population size $Seed$, Iterations $Step$, Crossover rate $Pc$, Mutation rate $Pm$;
		\ENSURE A vector of parameters ${\theta}^{\star}$;

		\STATE \textbf{Initialization} $Seed$, $Step$, $Pc$, $Pm$;

        \FOR{$seed$ $\leftarrow$ 0 to $Seed$}
            \STATE Encoding individual genotype ${G}_{seed}$;
        \ENDFOR

        \FOR{$steps$ $\leftarrow$ 0 to $Step$}
            \FOR{$seed$ $\leftarrow$ 0 to $Seed$}
                \STATE ${\theta}_{seed}(\mathcal{C}, {\mathcal{P}}_{l}, \mathcal{I}) \leftarrow {G}_{seed}$;
                \STATE ${X}_{adv}(seed)=S(X;{\theta}_{seed}(\mathcal{C}, {\mathcal{P}}_{l}, \mathcal{I}),\mathcal{M})$;
                \STATE ${f}_{Y}({X}_{adv}) \leftarrow f({X}_{adv})$;
                \IF{$f({X}_{adv}) \neq Y$}
                    \STATE ${\theta}^{\star}\leftarrow{\theta}_{seed}(\mathcal{C}, {\mathcal{P}}_{l}, \mathcal{I})$;
                    \IF{$\mathcal{T}({X}_{adv}) \rightarrow False$}
                        \STATE break;
                    \ENDIF
                    \STATE Output ${\theta}^{\star}$;
                    \STATE break;
                \ENDIF
            \ENDFOR
            \STATE Update: ${G}_{seed}\xleftarrow{Selection}{f}_{Y}({X}_{adv})$;
            \STATE Update: ${G}_{seed}\xleftarrow{Crossover} Pc$;
            \STATE Update: ${G}_{seed}\xleftarrow{Mutation} Pm$;

        \ENDFOR

	\end{algorithmic}  
\end{algorithm}

\subsection{Catoptric light adversarial attack}

AdvCL searches for $\theta(\mathcal{C}, {\mathcal{P}}_{l}, \mathcal{I})$, the physical parameters of catoptric light, to generate an adversarial sample ${X}_{adv}$ that deceives the classifier $f$. In this experiment, we consider a practical situation in which the attacker can only obtain the confidence score ${f}_{Y}(X)$ with given input image $X$ on ground truth label $Y$. In our proposed method, we use confidence score as the adversarial loss. Thus, the objective is formalized as minimizing the confidence score on the ground truth label $Y$, which can be expressed as:

\begin{equation}
    \label{Formula 3}
    \mathop{\arg\min}_{\theta}{\mathbb{E}}_{t \sim \mathcal{T}}[{f}_{Y}(t({X}_{adv}; \mathcal{C}, {\mathcal{P}}_{l}, \mathcal{I}))]
\end{equation}

$$s.t. \quad f({X}_{adv}) \neq Y$$


\begin{figure*}
\centering
\includegraphics[width=1\linewidth]{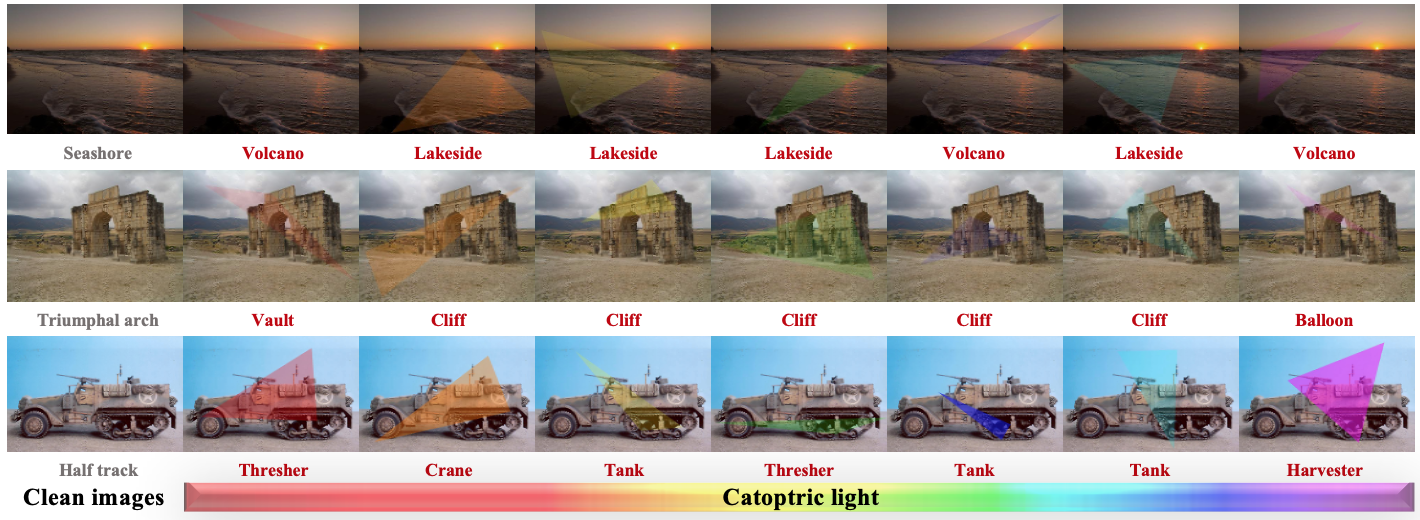} 
\caption{Digital samples generated by AdvCL.}.
\label{figure5}
\vspace{-3pt}
\end{figure*}

Algorithm \ref{algorithm1} shows the pseudocode of AdvCL. The proposed method takes clean image $X$, target classifier $f$, ground truth label $Y$, population size $Seed$, iteration number $Step$, crossover rate $Pc$, mutation rate $Pm$ as input decided by the attacker. Details of the algorithm are explained in Algorithm \ref{algorithm1}. In this case, we weed out the top tenth with the highest confidence score (the lower the confidence score, the more antagonistic) and then fill in the randomly encoded genes separately. The advantage of this selection strategy is that it saves time by directly eliminating the most incompetent individuals while also broadening the search scope and global optimization. Furthermore, the crossover rate $Pc$ and variation rate $Pm$ are set to 0.7 and 0.1, respectively. Finally, the algorithm outputs the physical parameter of the catoptric light ${\theta}^{\star}$, which is used in subsequent physical attacks.



\section{Evaluation}
\label{sec4}

\subsection{Experimental setting}

We put the proposed method to the test in both digital and physical settings. Consistent with AdvLB \cite{ref35}, we perform attacks using ResNet50 \cite{ref40} as the target model, and then randomly select 1000 ImageNet \cite{ref47} images that could be correctly classified by ResNet50 to perform digital tests. For physical examinations. Figure \ref{figure4} shows our experimental devices. We use iPhone 6s to take photos with color lamps as lighting equipment. Experiment verify that different camera devices would not affect the effectiveness of the proposed AdvCL. For all experiments, we use attack success rate (ASR) as a criterion to report the effectiveness of AdvCL, which is defined as follows:

\begin{equation}
\label{eq:Positional Encoding}
\begin{split}
    &{\rm ASR}(X) = 1-\frac{1}{N}\sum_{i=1}^{N}F({label}_{i})\\
    &F({label}_{i})=
        \begin{cases}
        1 & {label}_{i} \in {L}_{pre} \\
        0 & otherwise
        \end{cases}
\end{split}
\end{equation}

where $N$ is the number of clean samples in the dataset $X$, ${label}_{i}$ represents the ground truth label of the $i-th$ sample, ${L}_{pre}$  is the set of all labels predicted under attacking.

\subsection{Evaluation of effectiveness}
\textbf{Digital test.} On 1000 images that could be correctly classified by ResNet50, we test the effectiveness of AdvCL in digital environments. We draw three conclusions from the analysis of all experimental setups and results in Table \ref{Table 1}: \textbf{1)} The adversarial effect of AdvCL is better than that of the random attack method, demonstrating the feasibility and effectiveness of our method; \textbf{2)} The attack effectiveness increases as the number of edges ${\mathcal{P}}_{l}$ and intensity $\mathcal{I}$ increase, as expected; \textbf{3)} AdvCL achieves 87.9\% ASR under ${\mathcal{P}}_{l}$=3 and $\mathcal{I}$=0.6, demonstrating that our method is physically practical due to the legitimate configuration, subsequent physical tests follow this configuration. 
Figure \ref{figure5} shows some interesting results. The first column represents the samples we aim to attack, and the catoptric light of red, orange, yellow, green, blue, indigo and purple is added to clean samples respectively. The remaining columns are generated adversarial samples that can successfully fool the classifier. For example, when a blue ($\mathcal{C}(0,0,255)$) catoptric light is added to a clean sample, the seashore is misclassified as a volcano. On the other hand, when various catoptric lights are covered, the majority of the triumphal arch is misclassified as a cliff. Overall, AdvCL has an effective adversarial effect in the digital environment, causing advanced DNNs to misclassify target objects without changing their semantic information.

\begin{figure*}
\centering
\includegraphics[width=1\linewidth]{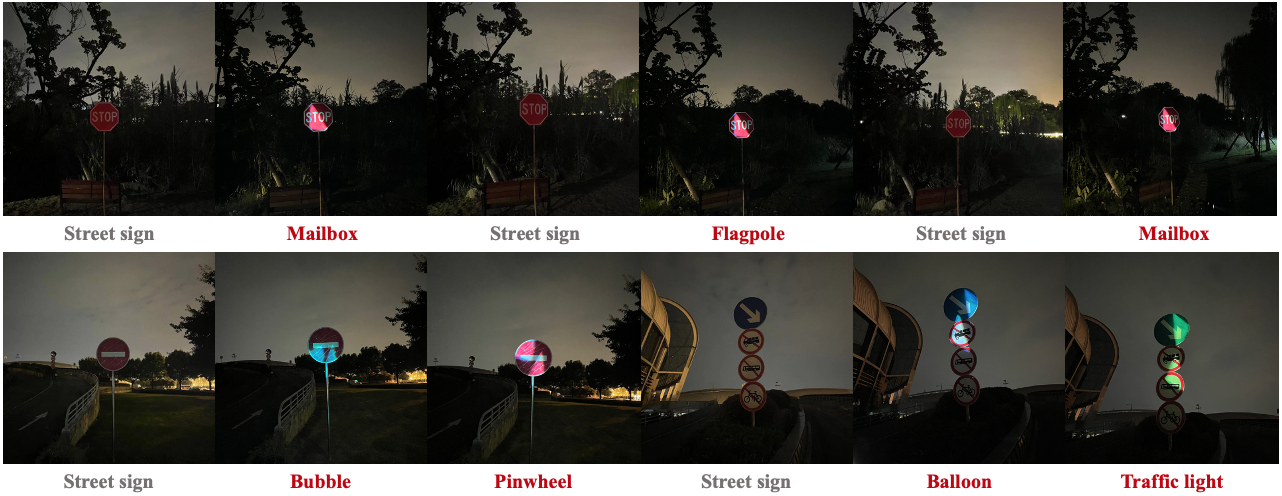} 
\caption{Physical samples generated by AdvCL.}.
\label{figure7}
\vspace{-3pt}
\end{figure*}

\textbf{Physical test.} We put the proposed method to the test in a physical setting. We use a strict experimental design in the physical test to demonstrate the rigor of AdvCL. Because environmental noise affects the robustness of physical attacks in the real world, we design indoor and outdoor tests separately. The indoor test eliminates the influence of outdoor noise, while the outdoor test reflects the performance of AdvCL in real-world scenarios.

As target objects for the indoor test, we use "Bonnet", "Plastic bag", "Street sign", and so on. and generate 30 adversarial samples with a 100\% ASR (ASR of 100\% in AdvLB \cite{ref35}). In the outdoor test, we choose "Street sign" as attack objects and generate 103 adversarial samples, achieving an ASR of 83.5\% (ASR of 77.4\% in AdvLB \cite{ref35}). Figure \ref{figure7} shows adversarial samples in an outdoor setting. The experimental results show that adding optimized catoptric light to the target objects causes advanced DNNs to misclassify the objects. To get closer to real-world scenarios, we conduct outdoor tests on a "Stop sign" from various angles, which demonstrates that AdvCL performs effective physical attacks on target objects from various angles.


\subsection{Evaluation of stealthiness}
As previously stated, we choose catoptric light as the physical perturbation in order to obtain a more natural physical sample, which makes our perturbation susceptible to being overlooked by human observers. As shown in Figure \ref{figure7}, our physical samples resemble natural catoptric light falling on a street sign, and human observers have difficulty distinguishing between natural and artificial catoptric light. On the other hand, the comparison of physical samples in Figure \ref{figure2} shows that the physical perturbation generated by AdvCL is more stealthy than the baseline. Given AdvCL's light-speed attack, AdvCL has greater temporal stealthiness than RP2 \cite{ref24} (RP2's physical perturbation will always adhere to the target object's surface, but AdvCL can control the light source, generating the physical perturbation only when the attack is carried out.). In contrast to AdvLB \cite{ref35}, the physical samples generated by AdvCL are more natural, allowing for better spatial stealthiness in our approach. When the cardboard is placed in front of the road sign for a shadow attack \cite{ref37}, it loses its spatial stealthiness, making human observers suspicious, whereas our method places the retroreflector far away from the target object, making AdvCL more stealthy than shadow attacks. In general, our approach results in a more stealthy attack than the baseline.

\subsection{Evaluation of robustness}

\textbf{Deploy AdvCL to attack various classifiers}.
We evaluate the robustness of the proposed AdvCL in a black-box setting with various classifiers, including the recently advanced DNNs (Inception v3 \cite{ref45}, VGG19 \cite{ref41}, ResNet101 \cite{ref40}, GoogleNet \cite{ref42}, AlexNet \cite{ref44}, MobileNet \cite{ref43}, DenseNet \cite{ref39}) and robust DNNs (Augmix+ResNet50 \cite{ref50}, ResNet50+RS \cite{ref51}, NF-ResNet50 \cite{ref52}). Note that the dataset is 1000 images selected from ImageNet that can be correctly classified by ResNet50. Table \ref{Table 2} shows the ASR of our method with different classifiers. AlexNet is found to be the most vulnerable in the black-box attack test, with a 97.2\% ASR and an average of 87.4 queries. Furthermore, robust DNNs such as Augmix+ResNet50, ResNet50+RS, and NF-ResNet50 are more robust. In general, the data in Table \ref{Table 2} show that AdvCL have an adversarial effect of ASR on various models by more than 80\% in the black-box setting, confirming the robustness of our proposed AdvCL.

\begin{table}
\centering
\setlength{\abovecaptionskip}{0.5cm}
\caption{\label{Table 2} Evaluation across various classifiers.}
\begin{tabular}{cccc}
\hline

\multicolumn{1}{c}{\multirow{2}{*}{$f$}} & \multicolumn{1}{c}{\multirow{1}{*}{\emph{w}/\emph{o} Attack}}

& \multicolumn{2}{c}{\emph{w}/ Attack}\\

\cmidrule(r){2-2}
\cmidrule(r){3-4}

\multicolumn{1}{c}{} & \multicolumn{1}{c}{Top-1 Accuracy} & ASR & Query \\
\hline

Inception v3 & 87.6 & 83.5 & 152.3 \\
\hline

VGG19 & 91.5 & 81.3 & 269.5 \\
\hline

ResNet101 & 96.1 & 82.1 & 198.4 \\
\hline

GoogleNet & 85.3 & 82.3 & 189.5 \\
\hline

AlexNet & 79.6 & 97.2 & 87.4 \\
\hline

MobileNet & 89.7 & 83.0 & 169.2 \\
\hline

DenseNet & 90.8 & 81.9 & 243.3 \\
\hline

Augmix+ResNet50 & 93.7 & 81.1 & 298.6 \\
\hline

ResNet50-RS & 94.6 & 80.9 & 330.4 \\
\hline

NF-ResNet50 & 94.8 & 80.3 & 346.1 \\
\hline

\end{tabular}
\end{table}

\begin{figure*}
\centering
\includegraphics[width=1\linewidth]{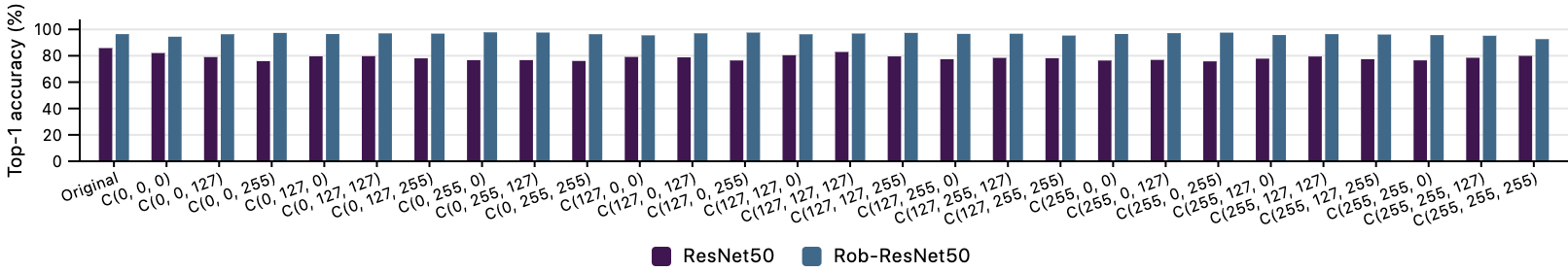} 
\caption{ResNet50 vs. Rob-ResNet50.}.
\vspace{-5mm}
\label{figure10}
\end{figure*}

\textbf{Transferability of AdvCL}.
Here, we demonstrate the attack transferability of AdvCL against advanced DNNs and robust DNNs in both digital and physical environments. As the dataset, we use the adversarial samples generated by AdvCL that successfully attacked ResNet50. Table \ref{Table 3} displays the experimental results. It can be seen that AdvCL has effective attack transferability in the digital environment, with an attack success rate of 77.24\% against AlexNet. In physical environments, AdvCL exhibits excellent attack transferability, with its transter attack paralyzing almost all advanced DNNs. 

Tables \ref{Table 2} and \ref{Table 3} show that our method is not only capable of performing black-box attacks against various classifiers, but also has a threatening adversarial effect on transfer attacks, which varifies the robustness of our proposed AdvCL.

\begin{table}
    \setlength{\abovecaptionskip}{0.cm}
    \setlength{\belowcaptionskip}{-0.cm}
    \centering
    \caption{\label{Table 3}Transferability of AdvCL (ASR(\%)).}
    \begin{tabular}{ccc}
    \hline
    $f$ & Digital & Physical \\
    \hline
    Inception v3 & 71.34 & 97.67\\
    \hline
    VGG19 & 55.85 & 100\\
    \hline
    ResNet101 & 47.21 & 97.67\\
    \hline
    GoogleNet & 67.02 & 100\\
    \hline
    AlexNet & 77.24 & 100\\
    \hline
    MobileNet & 63.01 & 97.67\\
    \hline
    DenseNet & 54.58 & 96.51\\
    \hline
    Augmix+ResNet50 & 52.67 & 84.35\\
    \hline
    ResNet50-RS & 50.91 & 81.65\\
    \hline
    NF-ResNet50 & 50.42 & 78.13\\
    \hline
    \end{tabular}
\end{table}

\section{Disccusion}
\label{sec5}




\subsection{Ablation study}
Here, we perform experiment to study the adversarial effects of Color $\mathcal{C}(r, g, b)$ on AdvCL. We selecte 27 colors of catoptric light to execute digital attacks on ResNet50, including $\mathcal{C}(0, 0, 0)$, $\mathcal{C}(0, 0, 127)$, $\mathcal{C}(0, 0, 255)$, $\mathcal{C}(0, 127, 0)$, $\mathcal{C}(0, 127, 127)$, $\mathcal{C}(0, 127, 255)$, $\mathcal{C}(0, 255, 0)$, $\mathcal{C}(0, 255, 127)$, $\mathcal{C}(0, 255, 255)$, $\mathcal{C}(127, 0, 0)$, $\mathcal{C}(127, 0, 127)$, $\mathcal{C}(127, 0, 255)$, $\mathcal{C}(127, 127, 0)$, $\mathcal{C}(127, 127, 127)$, $\mathcal{C}(127, 127, 255)$, $\mathcal{C}(127, 255, 0)$, $\mathcal{C}(127, 255, 127)$, $\mathcal{C}(127, 255, 255)$, $\mathcal{C}(255, 0, 0)$, $\mathcal{C}(255, 0, 127)$, $\mathcal{C}(255, 0, 255)$, $\mathcal{C}(255, 127, 0)$, $\mathcal{C}(255, 127, 127)$, $\mathcal{C}(255, 127, 255)$, $\mathcal{C}(255, 255, 0)$, $\mathcal{C}(255, 255, 127)$, $\mathcal{C}(255, 255, 255)$, All of the attacks achieve an ASR over 70\%,  among which $\mathcal{C}(255, 0, 255)$ achieves the highest ASR 96.60\%, and $\mathcal{C}(0, 0, 0)$ gets the lowest ASR 73.90\%. For a detailed experimental results see supplementary material.



\subsection{Defense of AdvCL}
In addition to demonstrating the potential threats of AdvCL, we attempt to defend against AdvCL with adversarial training. Here, in order to rigorously study the defense strategy against the proposed attack, we construct a larger dataset. To begin, we randomly select 50 clean samples from each of ImageNet's 1000 categories, yielding 50,000 clean samples. Second, add 27 color catoptric light to each clean sample with $\mathcal{I}$=0.5 to obtain the final data set, called ImageNet-CatoptricLight (ImageNet-CL), which contains 1.35 million adversarial samples.

For adversarial training, we use the proposed ImageNet-CL dataset. Torchvision is used to train the ResNet50 robust model (Rob-ResNet50). ADAM optimized the model on three 2080Ti GPUs with an initial learning rate of 0.01. Figure \ref{figure10} depicts the experimental results. Rob-ResNet50 achieves a classification accuracy of more than 90\% for adversarial samples. We deploy AdvCL to attack Rob-ResNet50 with ${\mathcal{P}}_{l}$=3, $\mathcal{I}$=0.6, and achieve 69.4\% ASR with 568.3 average queries (ResNet50: 87.9\% ASR, 184.9 average queries). This implies that while adversarial training can reduce AdvCL's ASR and increase AdvCL's attack time cost, it cannot completely defend against AdvCL. See the supplementary material for more experimental results on Rob-ResNet.



\section{Conclusion}
\label{sec6}

In this paper, we introduce AdvCL, a light-based physical attack that performs a physical attack by optimizing the physical parameters of catoptric light: ${\mathcal{P}}_{l}$, $\mathcal{C}(r, g, b)$, and $\mathcal{I}$. We take effectiveness, stealthiness, and robustness as criterion to evaluate the proposed method.
Extensive experimental design and results demonstrate the effectiveness of AdvCL in both digital and physical environments.
In terms of stealthiness, we present the generated physical sample and compare it to the baseline, demonstrating the proposed method's stealthiness in terms of both temporal and spatial stealthiness. We employ AdvCL to launch attacks on both advanced and robust DNNs. demonstrate the attack transferability of AdvCL, experimental results verified the robustness of our proposed AdvCL.
The proposed method reveals the physical-world security threat caused by light-based physical attacks. Our research also sheds new light on future physical attacks, such as using light as physical perturbations instead of stickers, which would improve physical attack flexibility. The proposed AdvCL is a useful complement to recent physical attacks as an effective, stealthy and robust light-based physical attack.

We will improve the proposed AdvCL to adapt to different tasks such as object detection, domain segmentation, etc. We will also focus on light-based physical attacks, such as adversarial projection, adversarial spot light, and so on. Furthermore, effective defense strategies against light-based attacks will emerge as a promising research area.

{\small
\bibliographystyle{ieee_fullname}
\bibliography{egbib}
}
\end{document}